\documentclass[review]{elsarticle}

\usepackage{lineno,hyperref}
\usepackage{amsmath}
\usepackage{multirow}
\usepackage{numcompress}
\usepackage{diagbox}
\usepackage[english]{babel}
\usepackage[T1]{fontenc}
\usepackage[utf8]{inputenc}
\usepackage{threeparttable}
\usepackage{booktabs} 
\newdefinition{definition}{Definition}
\newdefinition{theorem}{Theotem}
\newdefinition{proof}{Proof}
\modulolinenumbers[5]
\journal{Information Science}

\begin{document}

\begin{frontmatter}

\title{A Negation Quantum Decision Model to Predict the Interference Effect in Categorization}

\author[label1]{Qinyuan Wu}
\address[label1]{Institute of Fundamental and Frontier Science, University of Electronic Science and Technology of China, Chengdu, 610054, China}
\author[label1,label2,label3]{Yong Deng\corref{cor1}}
\ead{ dengentropy@uestc.edu.cn}
\cortext[cor1]{Corresponding author, Yong Deng, Institute of Fundamental and Frontier Science, University of Electronic Science and Technology of China, Chengdu, 610054, China}
\address[label2]{School of Education, Shannxi Normal University, Xi'an, 710062, China}
\address[label3]{School of Knowledge Science, Japan Advanced Institute of Science and Technology, Nomi, Ishikawa 923-1211, Japan}

\begin{abstract}
Categorization is a significant task in decision-making, which is a key part of human behavior. 
An interference effect is caused by categorization in some cases, which breaks the total probability principle. 
A negation quantum model (NQ model) is developed in this article to predict the interference. 
Taking the advantage of negation to bring more information in the distribution from a different perspective, the proposed model is a combination of the negation of a probability distribution and the quantum decision model. 
Information of the phase contained in quantum probability and the special calculation method to it can easily represented the interference effect. 
The results of the proposed NQ model is closely to the real experiment data and has less error than the existed models.
\end{abstract}
\begin{keyword}
  Categorization; interference effect; decision-making; quantum decision model; negation; quantum probability
\end{keyword}

\end{frontmatter}

\section{INTRODUCTION}
Decision making is one of the most significant parts of human behavior. 
Categorization has a important influence on a decision-making result \cite{clapper2017alignability,figueroa2016improving,yun2017utilizing}.
Some researchers have discovered that there is a classification fallacy in the categorization process.
In many decision-making behaviors, classification results are often an important basis for decision-making both for the human and some artificial intelligence systems.
Interference effects reflect the uncertainty information in the categorization has an important influence to the results.
Many theories have tried to find a theory and method that can quantify uncertainty,
for example, fuzzy set theory \cite{zimmermann2010fuzzy},  D-S evidence theory \cite{dempster2008upper,shafer1976mathematical}, Z number theory \cite{zadeh2011note}, etc.
 Decisions made after categorization do not conform to the full probability principle in the experiment \cite{busemeyer2009empirical,wang2016interference}
Townsend et.al proposed an experimental paradigm about categorization-decision process in 2000, and found that Markov decision process model could not predict this interference effect \cite{townsend2000exploring}.
The Markov model is a decision model based on classical probability. 
It still satisfies the classic full probability principle in essence, and cannot simulate the classification fallacy that breaks the full probability principle.
Classical probability cannot describe all the uncertain information in human being. 
There are some other new decision models to do it.
For example, a Markov decision model based on evidence theory \cite{he2018evidential}, quantum decision model based on quantum probability \cite{busemeyer2012quantum}, evidence quantum decision model combining evidence theory and quantum probability \cite{he2018evidential2} and so on. A new decision model from the perspective of negation and quantum probability to simulate and predict this classification interference effect is proposed in this article.
\par
The categorical disjunction fallacy which can be seen as a interference effect in the quantum decision model can be explained and predicted effectively.
Now, it is used in cognition science and decision-making very widly.
This type of model can successfully predict and explain the uncertainty information in a decision-making process. 
It can be used to explain many problems that are difficult to explain by traditional probability theories \cite{pothos2013can}, especially the decision-making behavior with uncertainty \cite{gronchi2017quantum}.
The model explains come contradictions in human decision-making such as the separation effect, the interference effect of classification decision, the combination fallacy, the average effect, the unboxing effect, the order effect of reasoning, and the prisoner’s dilemma \cite{tesavr2020quantum,chen2003quantum,busemeyer2011quantum,pothos2009quantum, khrennikov2009quantum}.
At present, tons of decision models based on the quantum probability have been proposed, 
be it the quantum cognitive model\cite{aerts2009quantum}, quantum-like models \cite{khrennikov2010ubiquitous}, generalized quantum models \cite{atmanspacher2002weak}, and so on. Compared with the traditional classification model, the difference and advantages of quantum decision model are mainly reflected in several aspects.
First, the quantum probability rule is different from the classical probability rule \cite{streater2000classical}. 
Born Law defines the probability of measuring a quantum state by the square of the amplitude \cite{born1926quantenmechanik}. 
Under the quantum probability system, since the state changes due to measurement, the quantum probability of transition between states is affected by all previous measurements \cite{zeh1970interpretation}.
In traditional models, such as Markov decision models, the latter state is only affected by the previous state \cite{puterman1990markov}.
Second, quantum probability does not satisfy Bell's inequality, which is a manifestation of its entanglement property \cite{cirel1980quantum}.
Quantum probability can reflect the entangled nature of things, which cannot be expressed by classical probability \cite{leggett1985quantum}.
Third, the order of quantum behavior observation cannot be exchanged. 
As far as quantum behavior is concerned, swapping the order of observation will result in different results. 
This trait also manifests itself in human behavior, which can be well simulated by quantum decision models \cite{wang2014context}.
The uncertainty relationship \cite{heisenberg1985anschaulichen} introduced by Heisenberg is a reflect of this characteristic.
In the actual decision-making process, the sequence of actions such as sorting or obtaining information will also affect the final decision-making result. 
With the help of the quantum theory, not only can the essential characteristics of how nature is organized be represented, but also a more accurate decision-making models is established \cite{wang2013potential}.
\par
Before humans make decisions, they often combine information from both forward and reverse thinking to make decisions. 
Thinking from another angle will result in some different information and get better answers.
The negation can bring higher-level information.
The essence of negation is to provide a quantitative method of negative information, thereby helping the system learn from mistakes.
Yager proposed a maximum entropy negation on the probability distribution \cite{yager2014maximum}.
In our previous work, an exponential negation for the probability distribution is proposed, which can converge to the maximum entropy distribution under any circumstances \cite{wu2020exponential}.
For a probability distribution, performing multiple iterations of the negation on it will increase its entropy and eventually converge to the maximum entropy state of the uniform distribution. 
In this process, we can use the negation to get higher-level information from the negative distribution.
This information is different from the information contained in the original distribution. 
To some extent, even if the initial information is wrong, negation and entropy can still help the system make correct decisions and knowledge management \cite{anjaria2020negation}.
In recent research, a new definition of $\chi$ entropy for complex probability distribution is proposed. Based on $\chi$ entropy, the author proposes a negation for complex probability distribution that can achieve the maximum $\chi$ entropy \cite{xiao2020maximum}. 
There are some quantum decision models that are improved in combination with evidence theory \cite{he2018evidential}, Bayesian quantum decision model combined with Bayesian network \cite{moreira2016quantum}.
We proposed a new negation quantum model (NQ model) which combine the exponential negation and the quantum decision model to use the new information in the negation to have better results.
\par 

This article has proposed a new negation quantum decision model to predict the inteference in categorization. 
We first define a quantum probability negation based on the classic probability exponential negation.
Then in each quantum decision-making stage, the quantum probability negation is used, and the quantum negative probability is used to represent the model state, to effectively use the high-order information provided by the negative probability in the quantum decision model.
In the end, the results obtained by this model are in good agreement with the original experimental data, and the interference effects of classification can be predicted
Compared with other existing models where the final data fluctuates near the original data, the data in this article has a certain bias and can better predict this phenomenon after correction.
\par
The organization structure of this article is expressed as follows: 
The basic of quantum decision models and the exponential negation will be introduced in Section \ref{PRLIMINARIES}.
The original classification decision experiment will be introduced in Section \ref{section3}.
Section \ref{section4} introduces the modeling proces s and details of the entire model.
The result of the model are presented in Section \ref{section5}.
Finally, the conclustion is in Section \ref{six}.

\section{PRLIMINARIES}
\label{PRLIMINARIES}
In this section, the quantum decision model \cite{busemeyer2012quantum} and the exponential negation \cite{wu2020exponential} are briefly introduced.
\subsection{Quantum decision model \cite{busemeyer2012quantum} }
The quantum decision model is a random swimming decision process \cite{busemeyer2012quantum} which is similar to the Markov model \cite{alagoz2010markov}.
However, the probability principle between the two methods is different. 
\par
Suppose a system has $n$ basic states which is a Hilbert space, every state can be described by the Dirac character as $|i\rangle, i=1,2,\dots,n$. 
All subsequent states of the system can be expressed linearly through these $n$ basic states. 
In the initial state (time $t=0$), these basic states can be represented by a wave function as $\psi_i (0), i=1,2,...,n$, 
then the state of a system can be expressed by a linearly combination of them as follows:
\begin{equation}
    \begin{split}
        |\Psi\rangle=\psi_1(0)|1\rangle+\psi_2(0)|2\rangle+\dots+\psi_n(0)|n\rangle=\sum\psi_i(0)|i\rangle
    \end{split}
\end{equation}
Use classical probability $p_i$ to represent the probability of state $|S_i\rangle$, the conversion relationship between the two is shown as follows:
\begin{equation}
    \begin{split}
    p_i=|\psi_i(t)|^2, i=1,2,\dots,n
    \end{split}
\end{equation}
The coordinate of the initial state in this Hilbert space can be presented as follows:
\begin{equation}
    \begin{split}
        \Psi(0)=\left[
 \begin{array}{c}
     \psi_1(0)  \\
     \psi_2(0) \\
     \dots \\
     \psi_n(0)
 \end{array}
 \right] 
    \end{split}
\end{equation}
The state will follow the Schrodinger equation and change over time:
\begin{equation}
\begin{split}
\label{Schrodinger}
    \frac{\mathrm{d}}{\mathrm{d}t}\Psi(t)=-iH\Psi(0)
\end{split}
\end{equation}
Where $H$ is a Hermitian matrix, $h_{ij}$ is the matrix element in the row $i$ and column $j$, which represents a transition rate from the state $|i\rangle$ to state $|j\rangle$. 
The relation of the different state during the time can be solved by the Iq \ref{Schrodinger} as follows:
\begin{equation}
    \begin{split}
        \Psi(t_1)=e^{-iHt}\Psi(t_0)=U(t)\Psi(0)
    \end{split}
\end{equation}
So at $t=t_1$, the state can be expressed as follows:
\begin{equation}
    \begin{split}
        |\Psi(t_1)\rangle=U(t)\sum\psi_{i}(0)|i\rangle=\sum\psi_i(0)U(t)|j\rangle=\sum\psi_i(0)\psi_i(t)
    \end{split}
\end{equation}
Then the conversion relation between quantum probability and classical probability is shown as follows:
\begin{equation}
\label{inteference}
    \begin{split}
        p=|\psi(t_1)|^2=|\sum_i\psi_i(0)\psi_i(t)|^2\ne\sum_i|\psi_i(0)|^2|\psi_i(t)|^2
    \end{split}
\end{equation}
It can be seen from Iq. (\ref{inteference})  that there is an interference effect after the categorization.
For the subsequent process, use $U(t)$ to represent the change of state, where $U(t)$ is a unitary matrix:
\begin{equation}
\label{move}
    \begin{split}
        U(t)=U(t)\Psi(0)
    \end{split}
\end{equation}
Iq. (\ref{move}) represents a dynamic process in the decision state.
A measurement operator can be set to take a measurement of states \cite{gardiner2004quantum}\cite{nielsen2002quantum}. 
$B$ is a selected operator which represents the states the categorization is $B$,
\begin{equation}
    \begin{split}
        B=\sum_{i>\theta}|j\rangle\langle j|
    \end{split}
\end{equation}
It is the experimenter who measured the system once and got the result of $B$. According to different results, the system state after the measurement was obtained.
\begin{equation}
    \begin{split}
        B*\psi=\left[
 \begin{array}{c}
     0  \\
     0 \\
     \psi_B 
 \end{array}
 \right] 
    \end{split}
\end{equation}
Similarly, when classified as $A$, the state can be represented as follows:
\begin{equation}
    \begin{split}
    A=\sum_{j<-\theta}|j\rangle\langle j| \\
        A*\psi=\left[
 \begin{array}{c}
     \psi_A  \\
     0 \\
     0 
 \end{array}
 \right] 
    \end{split}
\end{equation}
The measurement operator of the uncertain categorization $N$ is defined as,
\begin{equation}
\label{N}
    \begin{split}
    N=1-A-B\\
    N *\psi=\left[
 \begin{array}{c}
     0  \\
     \psi_N \\
     0 
 \end{array}
 \right]
    \end{split}
\end{equation}

\subsection{Exponential negation}
The exponential negation operation \cite{wu2020exponential} is an negation operation for the classical probability distribution. It is different from the negation operations on other probabilities. To some extent, it can be regarded as a geometric inverse operation. For any probability distribution, after multiple iterations, the negation operation can converge to the maximum entropy state, which shows that this negation operation can reveal all the high-order information of the distribution through multiple iterations.
\par
Suppose the event set is $A={A_1,A_2,\dots,A_n}$, the probability of event $A_i$ is $p_i$. The probability distribution is represented as $P={p_1,p_2,\dots,p_n}$ and the negation of the distribution is $\overline{P}={\overline{p_1},\overline{p_2},\dots,\overline{p_n}}$. The negation is expressed as the probability that the event $A_i$ will not occur. The definition of the negation is shown as follows:
\begin{equation}
    \begin{split}
        \overline{p_i}=(\sum_{i=1}^{n}e^{-p_i})^{-1}e^{-p_i}
    \end{split}
\end{equation}
\par
This exponential negation can still converge to the state of maximum entropy in every situation.
Many other negations of the probability distribution do not converge in the special binary case: $P={p_1=1, p_2=2}$.
Therefore, it is reasonable to choose this negation to establish the proposed negation quantum model.
\section{THE CATEGORIZATION EXPERIMENT}
\label{section3}
\subsection{Introduction of the experiment}
We use a paradigm proposed by Busemeyer et al \cite{busemeyer2009empirical} as a straightforward test for the negation quantum model. The classification decision experiment was designed by Townsend et al \cite{townsend2000exploring} to show that decisions are based on the classification information and not affected by the original information alone. It is first proposed to test the Markov decision model and then the paradigm was extended by Busemeyer \cite{busemeyer2009empirical} for the quantum model. 
\par
In this experiment, there were two different tested faces: the faces with 'narrow face' have narrow and wide lips and the faces with 'wide face' have wider faces and thin lips. 
Two actions are asked to do for the participants, 'attack' or 'withdraw'. A positive reward is given when the participants attack the 'bad guy'.
They are divided into two groups, and different group has a different condition.
The first group is to categorize the faces as belonging to either a 'good' guy or a 'bad' guy group then to decide to take which action. 
According to previous statistcs, participants are reminded that 60 percent of the narrow faces people are 'bas guy' and 60 percent of the wide one are 'good guy'.
The Busemeyer's paradigm \cite{busemeyer2009empirical} consists of two condition. 
For the first Categorization-Decision making paradigm (CD condition), the participants were asked to make a categorization first and then make an action decision.
An other condition is the Decision-Alone condition which means the participants only made an action decision without categorization.
Participants who attacked bad guys will be rewarded. 
Therefore, participants will always try to attack bad guys and avoid attacking good guys.
For both groups, the faces presented were the same. 
The sample size of the CD condition is $N=51\times26$, for 51 trails per person multiples 26 persons. 
The sample size of the D condition is  $N=17\times26=442$, for 17 trails per person multiples 26 persons.
\subsection{Experiment results}
\begin{table}[htbp]
\centering
\caption{The Busemeyer's results of the CD condition and D-Alone condition \cite{busemeyer2009empirical}}
\begin{tabular}{cccccccc}
\midrule
Face Type & $P(G)$ & $P(A|G)$  & $P(B)$  & $P(A|B)$ & $P_t$ &$P(A)$ &t\\
\hline
      W(Wide)       & 0.84 & 0.35 & 0.16 & 0.52 & 0.37 &0.39 &0.5733 \\
      N(Narrow)     & 0.17 & 0.41 & 0.83 & 0.63 & 0.59 &0.69 &2.54 \\
\bottomrule
\end{tabular}
\label{ExResult}
\end{table}
Table \ref{ExResult} shows the experiment results. 
The column labeled $P(G)$ shows the probability to classify the face as 'good guy'. 
The column labeled $P(A|G)$ shows the conditional probability of choosing an attack action under the condition of being classified as a 'good'. 
The column labeled $P(B)$ shows that the probability to classify the face as 'bad guy'. 
The column labeled $P(A|B)$ shows the probability of choosing an attack action under the condition of being classified as a 'bad guy'. 
The column labeled $P_t$ represents the probability which is calculate as follows:
\begin{equation}
    \begin{split}
        P(G)P(A|G)+P(B)P(A|B)=P_t
    \end{split}
\end{equation}
The column labeled $P(A)$ represented the probability of choosing action 'attack' during the D condition. The column labeled $t$ is the t-test value.
\par
$P_t=P(A)$ is easy to get based on the law of total probability and the sure thing principle.
However, the result shows that $P_t\ne P(A)$ which means there exist some interference effects during the C-D condition, especially for the narrow faces. 
The t-test value is calculated to indicates the level of significance of the difference between $P_t$ and $P(A)$ \cite{busemeyer2009empirical}. The t-test value shows that the difference is statically significant in the narrow-face data, however, is not statically significant in the wide-face data. In this article, only the narrow-face data is considered to use to verify the validity of the proposed model.
\begin{table}[htbp!]
\label{data}
\begin{threeparttable} 
\centering
\caption{All results of categorization decision-making experiments }
\begin{tabular}{cccccccc}
\midrule
Literature & Face Type & $P(G)$ & $P(A|G)$  & $P(B)$  & $P(A|B)$ & $P_t$ &$P(A)$ \\
\hline
Busemeyer et al. \cite{busemeyer2009empirical} &      W   & 0.84 & 0.35 & 0.16 & 0.52 & 0.37 &0.39 \\
  &    N   & 0.17 & 0.41 & 0.83 & 0.63 & 0.59 &0.69 \\
  Wang and Busemeyer \cite{wang2016interference} & & & & & & & \\
  Experiment 1 & W& 0.78&0.39 &0.22& 0.52&0.42&0.42\\
   &N&0.21&0.41&0.79&0.58&0.54&0.59\\
   Experiment 2 & W& 0.78&0.33 &0.22& 0.53&0.37&0.37\\
   &N&0.24&0.37&0.76&0.61&0.55&0.60\\
   Experiment 3(a) & W& 0.77&0.34 &0.23& 0.58&0.40&0.39\\
   &N&0.24&0.33&0.76&0.66&0.58&0.62\\
   Experiment 3(b) & W& 0.77&0.23 &0.23& 0.69&0.34&0.33\\
   &N&0.25&0.26&0.75&0.75&0.63&0.64\\
   Average & W& 0.79&0.33 &0.21& 0.57&0.38&0.38\\
   &N&0.22&0.36&0.78&0.65&0.58&0.63\\
\bottomrule
\end{tabular}
\label{AllResult}
\begin{tablenotes}   
\footnotesize              
\item[1] In the Busemeyer's experiment \cite{busemeyer2009empirical}, presents the most classic classification decision experiment results.       
\item[2] In the Wang and Busemeyer's experiment \cite{wang2016interference}, the experiment 1 replicate the classic experiment with a larger data set (N=721). Experiment 2 conducted a X-D condition and a C-D condition, in this article, we only use the C-D condition's data. Experiment 3 tried different reward amounts. In experiment 3(a), the reward for attacking bad guys is less than the reward for experiments 1 and 2. In experiment 3(b), the reward for attacking bad guys is more than the reward for experiments 1 and 2.
\end{tablenotes}           
\end{threeparttable}
\end{table}
Some other works also have analysed this paradigm. 
Table \ref{AllResult} shows all the results in these experiments. 
In these experiments, the classification brings the interference effect, affects the decision result and breaks the total probability law. 
\section{METHOD}
\label{section4}
\begin{figure}[!htbp]
  \centering
  \includegraphics[scale=0.4]{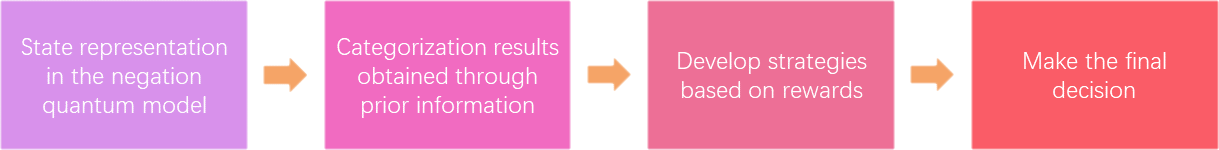}
  \caption{The four steps of model construction}
  \label{Flow}
\end{figure}

The negation quantum model is designed to be four main parts, the overall framework is shown in Figure \ref{Flow}.
\subsection{State representation under the negation quantum framework}
In the experiment mentioned before, for the C-D condition, a participant had two choices in the classification section, 
choose 'good guy'(abbreviated as G) or 'bad guy'(abbreviated as B), and then choose to 'attack'(abbreviated as A) or 'withdraw'(abbreviated as W). 
Considering that human decision making is not always deterministic, an action representing uncertain information can be added to improve the accuracy. 
The proposed model add an uncertain action to the action set which is 'hesitate'(abbreviated as H). 
For a categorization decision-making process, there are a total of six basic states, the Dirac symbol is used here to represent a state.
\begin{equation}
    \nonumber
    \begin{split}
    |AG\rangle, |HG\rangle, |WG\rangle, |AB\rangle, |HB\rangle, |WB\rangle
    \end{split}
\end{equation}
State $|AG\rangle $ represents the state of classifying the face as G but still choose attacking. 
State $|HG\rangle$ represents the state of classifying the face as G and hesitating to attack. 
The six states span a Hilbert space which contains all the states in which people may make choices.
Any state in this space can be expressed as a linear combination of these six states:
\begin{equation}
\label{state}
    \begin{split}
        |\Psi\rangle &= \psi_{AG}|AG\rangle+\psi_{HG}|HG\rangle+\psi_{WG}|WG\rangle 
        +\psi_{AB}|AB\rangle+\psi_{HB}|HB\rangle+\psi_{WB}|WB\rangle\\
        &=\left[
        \begin{array}{cccccc}
            \psi_{AG}&\psi_{HG}&\psi_{WG}&\psi_{AB}&\psi_{HG}&\psi_{WG}
        \end{array}
        \right]\left[
        \begin{array}{c}
             |AG\rangle  \\
             |HG\rangle  \\
             |WG\rangle  \\
             |AB\rangle  \\
             |HB\rangle  \\ 
             |WB\rangle  
        \end{array}
        \right]
    \end{split}
\end{equation}
 $|~ \rangle$ is the Dirac notation, $|AG\rangle$ represented a state that the participant classified a face as G but choose to attack. 
 $\psi_{AG}$ is a wave function or the quantum probability and $|\psi_{AG}\rangle ^2$ represents the probability of state $|AG\rangle$. 
 $[~\psi_{AG}~\psi_{HG}~\psi_{WG}~\psi_{AB}~\psi_{HG}~\psi_{WG}~]$ can be seen as the  coordinate for state $|\Psi\rangle$ in the space.
 \par
 The negation of the quantum probability is defined by the exponential negation of the probability distribution \cite{wu2020exponential}.
\begin{definition}
\label{negation}
Suppose there is a quantum probability distribution as $\Psi=\{\psi_1, \psi_2, \dots, \psi_n\}$,
the corresponding probability distribution is $P=\{|\psi_1|^2, |\psi_2|^2, \dots, |\psi_n|^2\}$, the probability distribution follows the probability principle as $\sum_{i=1}^n p_i=1$
the negation of the quantum probability distribution is defined as follows:
\begin{equation}
    \begin{split}
        \overline{\psi_{ij}}=(\sum_i^n e^{-|\psi_{i}|^2})^{\frac{1}{2}}e^{-\frac{1}{2}|\psi_{i}|^2}
    \end{split}
\end{equation}
\end{definition}\par
The quantum probability under Definition \ref{negation}  satisfies the normalization condition, $\sum_{i=1}^n|\overline{\psi_{i}}|^2=1$. 
A negation of quantum probability corresponds to multiple quantum probabilities with different phases for the same magnitude of the modulus, which means that Definition \ref{negation} can be regarded as a representation of a set of multiple quantum negative probabilities.\par
$\overline{\psi_{i}}$ represents the quantum probability that state $i$ will not occur, for example, $|\overline{\psi_{AB}}\rangle$ can represent the quantum probability of being classified as a bad person and then attacking. So, quantum probability can be represented by inverse quantum probability as follows:
\begin{equation}
    \begin{split}
        \psi_{AB}\longleftarrow \frac{1}{5}(\overline{\psi_{AG}}+\overline{\psi_{HG}}+\overline{\psi_{WG}}+\overline{\psi_{HB}}+\overline{\psi_{WB}})
    \end{split}
\end{equation}
\par
For a C-D condition, only the six basic states will appear. As the $\overline{\psi_{ij}}$ represents the probability of not appearing of state $ij$, if the other five states do not appear, then the remaining one state will appear. So the quantum probability of state $AB$ can be represented by the negation quantum probability of the other five states. In the initial state, the probability of all states is the same. Here, the average value is taken for each state.
\par
The state coordinate in Eq. (\ref{state}) can be expressed by the negation of quantum probability as follows:
\begin{equation}
    \label{negation_state}
    \begin{split}
        |\Psi_{t_0}\rangle=\frac{1}{5}\left[
            \begin{array}{c}
                \overline{\psi_{HG}}+\overline{\psi_{WG}}+\overline{\psi_{WG}}+\overline{\psi_{HB}}+\overline{\psi_{WB}}\\
                \overline{\psi_{AG}}+\overline{\psi_{WG}}+\overline{\psi_{WG}}+\overline{\psi_{HB}}+\overline{\psi_{WB}}\\
                \overline{\psi_{AG}}+\overline{\psi_{HG}}+\overline{\psi_{WG}}+\overline{\psi_{HB}}+\overline{\psi_{WB}}\\
                \overline{\psi_{AG}}+\overline{\psi_{HG}}+\overline{\psi_{WG}}+\overline{\psi_{HB}}+\overline{\psi_{WB}}\\
                \overline{\psi_{AG}}+\overline{\psi_{HG}}+\overline{\psi_{WG}}+\overline{\psi_{AB}}+\overline{\psi_{WB}}\\
                \overline{\psi_{AG}}+\overline{\psi_{HG}}+\overline{\psi_{WG}}+\overline{\psi_{AB}}+\overline{\psi_{HB}}
            \end{array}
        \right]^T\left[
        \begin{array}{c}
             |AG\rangle  \\
             |HG\rangle  \\
             |WG\rangle  \\
             |AB\rangle  \\
             |HB\rangle  \\ 
             |WB\rangle  
        \end{array}
        \right]=\Psi(0)\left[
            \begin{array}{c}
                 |AG\rangle  \\
                 |HG\rangle  \\
                 |WG\rangle  \\
                 |AB\rangle  \\
                 |HB\rangle  \\ 
                 |WB\rangle  
            \end{array}
            \right]
    \end{split}
\end{equation}
\par
Denote $\Psi(0)$ as the coordinate for the initial state $|\Psi_0\rangle$, for convenience later, $\Psi(0)$ can be expressed as follows:
\begin{equation}
    \begin{split}
    \Psi(0)=
        \frac{1}{5}\left[
        \begin{array}{c}
             B-\overline{\psi_{AG}}  \\
             B-\overline{\psi_{HG}}\\
             B-\overline{\psi_{WG}}\\
             B-\overline{\psi_{AB}}\\
             B-\overline{\psi_{HB}}\\
             B-\overline{\psi_{WB}}
        \end{array}
        \right]^T
    \end{split}
\end{equation}
Where $B=\overline{\psi_{AG}}+\overline{\psi_{HG}}+\overline{\psi_{WG}}+\overline{\psi_{AB}}+\overline{\psi_{HB}}+\overline{\psi_{WB}}$.\par
The same state representation is used for both the C-D and D condition. The interference occurs during the categorization process, the model will distinguish between the two situations in the second step.\par
\subsection{Categorization results obtained through prior information}
\begin{figure}[!htbp]
  \centering
  \includegraphics[scale=0.5]{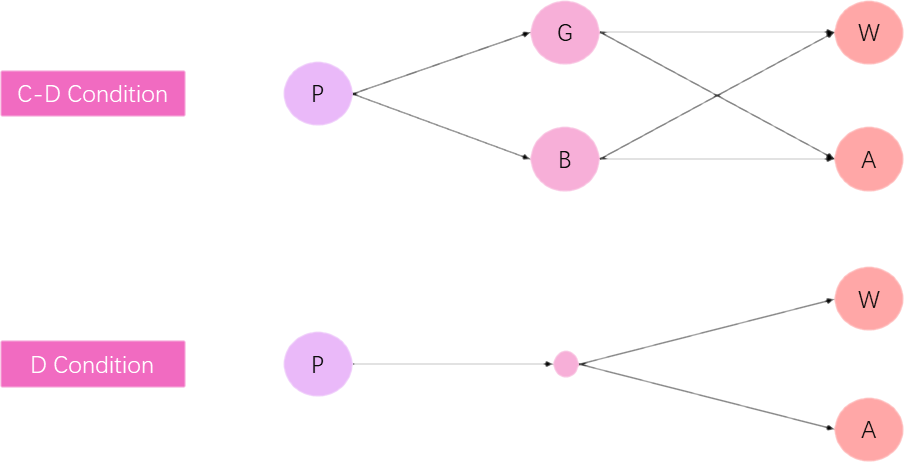}
  \caption{The decision-making progress for C-D condition and D condition}
  \label{interference}
\end{figure}

For the influence of the interference phenomenon caused by classification on the final decision, we can use the double-slit diffraction experiment in physics to understand. 
As shown in Figure \ref{interference},  for the C-D condition, the participant made the final decision through a categorization process,  the final decision is made based on the two classification results. 
Like in a double-seam experiment, electron waves interfere as they pass through two seams because the path to the final screen has changed, the probability wave has a phase difference.
The interference occurs in the C-D condition because the categorization caused two different paths to the final decision.
For the D condition, the participant make a decision directly. 
\par 
It is considered that the participant has performed a measurement on the system and obtained the result of $G$ and $B$. 
According to the different results, the measured system state is obtained.
Here we use the classification information to establish the operator of measurement.
For the C-D condition, the participant was told that there are 60 percent of the narrow face people are 'bad guy'. 
According to this prior information, the state changed to a new state $\Psi_{t_1}$ at the time of $t_1$, the coordinate can be denoted as $\Psi(1)$. 
For between good and bad, get a different quantum probability distribution. 
When the state changes, the change in the distribution is defined as follows for the faces which are categorized as G('good guy').
\begin{equation}
    \begin{split}
        \Psi(1)=\frac{1}{\sqrt{P_G}}\left[
            \begin{array}{c}
                B-\overline{\psi_{AG}}\\
                B-\overline{\psi_{HG}}\\
                B-\overline{\psi_{WG}}\\
                0\\
                0\\
                0
            \end{array}
        \right]=\left[
            \begin{array}{c}
                \Psi_G^*\\
                0
            \end{array}
        \right]
    \end{split}
\end{equation}
\par
Where $P_G=\psi_{AG}^2+\psi_{HG}^2+\psi_{WG}^2$ and $\Psi_G^*$ is a $3\times 1$ matrix which represents the probability distribution of actions selected based on the categorization as G.
Therefore, $\Psi_G^*$ needs to be normalized to be $\Psi_G$ which gets $||\Psi_G||_2=1$.
\begin{equation}
    \begin{split}
        \Psi_G=\frac{\Psi_G^*}{\sqrt{||\Psi_G^*||_2}}
    \end{split}
\end{equation}
If the face is categorized as B('bad guy'), the distribution across states changes as follows:
\begin{equation}
    \begin{split}
        \Psi(1)=\frac{1}{\sqrt{P_B}}\left[
            \begin{array}{c}
                0\\
                0\\
                0\\
                B-\overline{\psi_{AB}}\\
                B-\overline{\psi_{HB}}\\
                B-\overline{\psi_{WB}}
            \end{array}
        \right]=\left[
            \begin{array}{c}
                \Psi_B^*\\
                0
            \end{array}
        \right]
    \end{split}
\end{equation}
\begin{equation}
    \begin{split}
        \Psi_B=\frac{\Psi_B^*}{\sqrt{||\Psi_B^*||_2}}
    \end{split}
\end{equation}
Where $P_B=1-P_G$ and $\Psi_B$ is a $3\times 1$ matrix which represents the probability distribution of actions selected based on the categorization as B.
\par 
For the D condition, there is no categorization during this time, the state didn't have any different with the initial one.
\begin{equation}
    \label{D condition t1}
    \begin{split}
        \Psi(1)=\Psi(0)=\left[
            \begin{array}{c}
                P_G \Psi_G^*\\
                P_B \Psi_B^*
            \end{array}
        \right]=P_G \sqrt{||\Psi_G^*||_2}\left[
            \begin{array}{c}
                \Psi_G\\
                0
            \end{array}
        \right]+P_B \sqrt{||\Psi_B^*||}_2\left[
            \begin{array}{c}
                0\\
                \Psi_B
            \end{array}
        \right]
    \end{split}
\end{equation}
\par
Eq. (\ref{D condition t1}) indicates that the distribution of D condition at time $t_1$ can be seen as a weighted sum of the G categorization and the B categorization.

\subsection{Develop strategies based on rewards}
For the participants, they must choose the action that is most beneficial to them. Rewards are used to measure how much benefit a chosen action brings to the participate.
After selecting an action at time $t_2$, state $|\Psi(t_1)\rangle$ changes into state $|\Psi(t_2)\rangle$, the coordinate can be denoted as $\Psi(2)$.
Based on the quantum dynamical modeling \cite{busemeyer2011quantum}, this state transition leads to the decision making and the state transition obeys the Schrodinger equation (Eq. (\ref{Schrodinger})).
\begin{equation}
    \begin{split}
        \Psi(2)=e^{iHt}\Psi(1)
    \end{split}
\end{equation}
Where $H$ is a Hamiltonian matrix:
\begin{equation}
\begin{split}
    H=\left[
        \begin{array}{cc}
            H_G&0\\
            0&H_B
        \end{array}
    \right]
\end{split}
\end{equation}
$H_G$ is applied for the categorization of good guys and $H_B$ is applied for the categorization of bad guys.
Both $H_G$ and $H_B$ are $3\times 3$ matrixes as a function of the difference between different actions which are defined as follows:
\begin{equation}
    \begin{split}
        H_G=\frac{1}{1+h_G^2}\left[
            \begin{array}{ccc}
                h_G&0&1\\
                0&1&0\\
                1&0&-h_G
            \end{array}\right],
        H_B=\frac{1}{1+h_B^2}\left[
            \begin{array}{ccc}
                h_B&0&1\\
                0&1&0\\
                1&0&-h_B
             \end{array}\right]
    \end{split}
\end{equation}
Where $h_G$ and $h_B$ are two parameters which can be seen as a reward function for the certain action. These two parameters play a critical role in this state transition.
\par 
From above, the state transition equation from $t_1$ to $t_2$ for the C-D condition is as follws:
\begin{equation}
\label{Psi2_CD}
    \begin{split}
        \Psi(2)_G=e^{iHt}\Psi(1)=\left[
            \begin{array}{cc}
                e^{-iH_Gt}&0\\
                0&e^{-iH_Bt}
            \end{array}
        \right]\left[
            \begin{array}{c}
                \Psi_G\\
                0
            \end{array}
        \right]=e^{-iH_Gt}\Psi_G, ~Categorization~as~G\\
        \Psi(2)_B=e^{iHt}\Psi(1)=\left[
            \begin{array}{cc}
                e^{-iH_Gt}&0\\
                0&e^{-iH_Bt}
            \end{array}
        \right]\left[
            \begin{array}{c}
                0\\
                \Psi_B
            \end{array}
        \right]=e^{-iH_Bt}\Psi_B, ~Categorization~as~B
    \end{split}
\end{equation}
For the D condition, the equation is as follows:
\begin{equation}
    \label{D decision}
    \begin{split}
        \Psi(2)&=e^{iHt}\Psi(0)=\left[
            \begin{array}{cc}
                e^{-iH_Gt}&0\\
                0&e^{-iH_Bt}
            \end{array}
        \right]\left[
            \begin{array}{c}
                P_G\sqrt{||\Psi_G^*||}_2\Psi_G\\
                P_B\sqrt{||\Psi_B^*||}_2\Psi_B
            \end{array}
        \right]\\
        &=\sqrt{P_G}e^{-iH_Gt}\sqrt{||\Psi_G^*||}_2\Psi_G+\sqrt{P_B}e^{-iH_Bt}\sqrt{||\Psi_B^*||}_2\Psi_B
    \end{split}
\end{equation}
\par 
Although the decision making progress of D condition is an unknown progress,  according to Eq. (\ref{D condition t1}) and Eq. (\ref{D decision}), it can always be represented as a weighted sum of the two known results in the C-D condition.
\subsection{Make the final decision}
At time $t_2$,  the quantum probability distribution for each of the basic states before the final decision is made.
But the distribution at this point contains the hesitating items which is not a allowed action. 
Here we use the idea of evidence fusion in evidence theory, the mass function and a measure matrix which developed a linear combination rule are used to assign the probability of the hesitancy action to the two allowed action.
Firstly, the measure matrix is used to get the conditional mass function , $m(A|G)$ for example, which means the 'probability' of the action 'attack' when the face is classified as G under the evidence theory framework. And then, the real probability of the action can be calculated based on the mass function.\par 
The measure matrix is defined as follows:
\begin{equation}
    \begin{split}
        M=\left[
            \begin{array}{cc}
                M_G&0\\
                0&M_B
            \end{array}
        \right]
    \end{split}
\end{equation}
\par
Where $M_G$ and $M_B$ are both $3\times 3$ matrix to pick out a certain action from the action amplitude distribution Eq. (\ref{Psi2_CD}). $M_G$ is for the conditional distribution classified as G, $M_B$ is for the conditional distribution classified as B. \par
For the C-D condition, to get the mass function of withdraw action, $M_{WG}$ and $M_{WB}$ are defined as follows:
\begin{equation}
    \begin{split}
        M_{WG}=M_{WB}=diag(0, 0, 1)
    \end{split}
\end{equation}\par
The mass function can be calculated as follows:
\begin{equation}
    \begin{split}
        m(W|G)&=||M_W \Psi_G(2)||^2=\left[\begin{array}{cc}
            M_{WG} & 0 \\
            0 & M_{WB}
        \end{array}\right]\left[
        \begin{array}{cc}
            e^{-itH_G} &0  \\
            0 &e^{-itH_B} 
        \end{array}
        \right]\left[
        \begin{array}{c}
             \Psi_G \\
             0 
        \end{array}
        \right]
        =M_{WG}e^{-itH_G}\Psi_G\\
         m(W|B)&=||M_B \Psi_B(2)||^2=\left[\begin{array}{cc}
            M_{WG} & 0 \\
            0 & M_{WB}
        \end{array}\right]\left[
        \begin{array}{cc}
            e^{-itH_G} &0  \\
            0 &e^{-itH_B} 
        \end{array}
        \right]\left[
        \begin{array}{c}
             0 \\
             \Psi_B 
        \end{array}
        \right]
        =M_{WB}e^{-itH_B}\Psi_B
    \end{split}
\end{equation}\par
The matrix of hesitancy action, $M_HG$ and $M_HB$ can be defined as follows:
\begin{equation}
    \begin{split}
    M_{HG}=M_{BG}=diag(0,1,0)
    \end{split}
\end{equation}\par
The mass function can be calculated as follows:
\begin{equation}
    \begin{split}
        m(H|G)&=||M_H \Psi_G(2)||^2=\left[\begin{array}{cc}
            M_{HG} & 0 \\
            0 & M_{HB}
        \end{array}\right]\left[
        \begin{array}{cc}
            e^{-itH_G} &0  \\
            0 &e^{-itH_B} 
        \end{array}
        \right]\left[
        \begin{array}{c}
             \Psi_G \\
             0 
        \end{array}
        \right]
        =M_{HG}e^{-itH_G}\Psi_G\\
         m(H|B)&=||M_B \Psi_B(2)||^2=\left[\begin{array}{cc}
            M_{HG} & 0 \\
            0 & M_{HB}
        \end{array}\right]\left[
        \begin{array}{cc}
            e^{-itH_G} &0  \\
            0 &e^{-itH_B} 
        \end{array}
        \right]\left[
        \begin{array}{c}
             0 \\
             \Psi_B 
        \end{array}
        \right]
        =M_{HB}e^{-itH_B}\Psi_B
    \end{split}
\end{equation}
\par 
In order to get the conditional probability of a certain action 'attack' and 'withdraw', the 'probability' (mass function) of the uncertain action, hesitancy, is divided into two parts for the action 'attack' and action 'withdraw'. In order to have less parameters in this model, the probability of choosing two actions is equal when the participate hesitate. Considering there are only two certain actions, a participate doesn't choose 'withdraw' means that 'attack' is chosen. Therefore, the model also use a negation of the quantum probability distribution to calculate. The calculate principle is defined as follows:
\begin{equation}
    \begin{split}
        P(A|G)=||\overline{\Psi(W|G)}+\frac{1}{2}\overline{\Psi(H|G)}||^2
    \end{split}
\end{equation}
Where $\Psi(W|G)=\sqrt{m(W|G)}$, $\overline{\Psi(W|G)}$ and $\overline{\Psi(W|G)}$ follows the calculation method of Definition \ref{negation}.\par
After getting the conditional probability of choosing 'attack' when the face is classified as G, the new information is got. 
Considering that when a participate makes a decision, the probability of choosing 'attack' when the face is categorized as G is a negation of choosing 'attack' when the face is categorized as B, 
the proposed model uses the exponential negation to get the negation probability.
\begin{equation}
    \begin{split}
        P(A|B)=\overline{P(A|G)}=(e^{-P(A|G)}+e^{-(P(A|G)+\alpha)})^{-1}e^{-P(A|G)}
    \end{split}
\end{equation}
Where $\alpha$ is a parameter which means the sum of $P(A|B)$ and $P(A|G)$.
\par
The final attack probability is calculated based on the total probability principle as follows:
\begin{equation}
\label{CD_result}
    \begin{split}
        P(A)=P(G)P(A|G)+P(B)P(A|B)
    \end{split}
\end{equation}\par 
For the D condition, since there is no categorization process, the extraction of the certain actions is done together with the uncertain action, the measure matrix is defined as follows:
\begin{equation}
    \begin{split}
        M_G=M_B=diag(1, \frac{1}{\sqrt{2}}, 0)
    \end{split}
\end{equation}
The probability of 'attack' action is calculated as follows:
\begin{equation}
\label{D_result}
    \begin{split}
        P(A)&=||M e^{-itH}\Psi(0)||^2\\
        &=||P(G)M_Ge^{-iH_Gt}\Psi_G+P(B)M_Be^{-iH_Bt}\Psi_B||^2
    \end{split}
\end{equation}
\par 
The final results of the two conditions are different. The difference reflects the fact that the categorization caused a interference to the decision-making and the interference caused the difference of the results.
For the C-D condition, the final decision of the decision-making is made through two different paths, Eq. (\ref{CD_result}) presents the form of adding two parts which shows that the two paths both have a influence to the final decision. For the D condition, there is only one path from the initial state to the final decision state, Eq. (\ref{D decision}) models the progress as a 'weighted sum' of the C-D's two paths.


\section{RESULT}
\label{section5}
\subsection{Determine the parameter}
In the proposed model, the time parameter $t$ is set as $t=\frac{\pi}{2}$, which is same as the prior quantum model \cite{busemeyer2009empirical}\cite{busemeyer2012quantum}\cite{he2018evidential}. 
The proposed NQ model has four free parameters, $h_G, h_B$, $\alpha$ and $P_G$, 
which can be determined by minimizing the square error of the experiment variables. 
\begin{equation}
\label{optimization}
    \begin{split}
        arg~min_{\alpha, h_G, h_B}\sqrt{\sum_{i=1}^{6}(X_i-X_0)^2}
    \end{split}
\end{equation}
Where $ X_1=P(G),X_2=P(A|G),X_3=P(B), X_4=P(A|B), X_5=P_t, X_6=P(A)$, $X_{i0}$ is the real data from the experiment which is shown in Table. 2. 
 The initial quantum probability distribution is defined as a uniform distribution.
By solving Eq. (\ref{optimization}), all the free parameters can be determined.
\subsection{Results analyse}
The result calculated by the proposed NQ model is shown in Table 3. The initial results show a certain overall bias, however, it is easy to modify as follows:
\begin{equation}
    \begin{split}
        E_0&=R_0-I\\
        R&=R_0-\frac{1}{N}\sum_{i=1}^NE_{0i}
    \end{split}
\end{equation}
Where $R_0$ is the initial results from the NQ model, $I$ is the initial data from the experiments, $E_0$ is the initial error between them. $R$ is the modified results by subtract the mean of the initial error. After this simple modify, we can get a result that is in great agreement with the experimental data.\par
\begin{figure}[!htbp]
  \centering
  \includegraphics[scale=0.7]{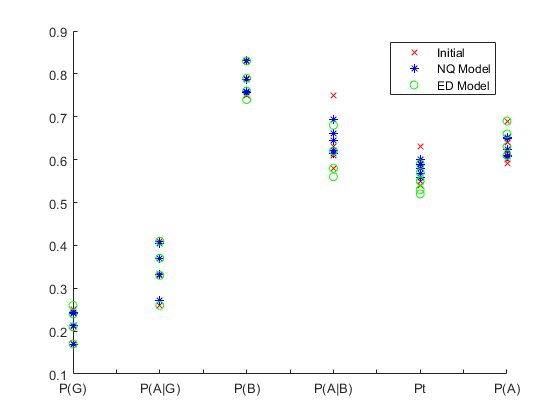}
  \caption{Comparison of experimental data with results of two models}
  \label{compare_result}
\end{figure}

\begin{figure}[!htbp]
  \centering
  \includegraphics[scale=0.7]{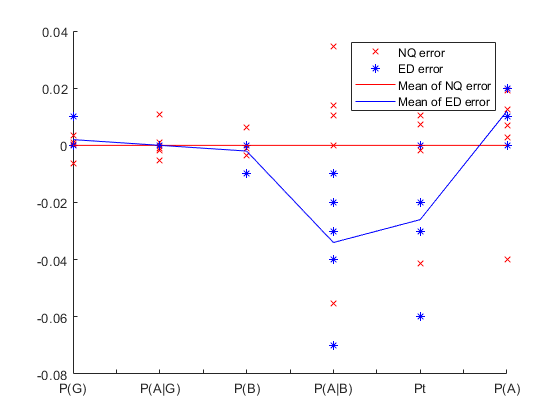}
  \caption{Comparison of the error of two models}
  \label{compare_error}
\end{figure}
An prior evidential dynamical model \cite{he2018evidential} (ED model) which combines the evidence theory and the quantum model shows a great result to predict the interference effect in this decision-making experiment. Figure \ref{compare_result} shows a comparison of the results of the proposed NQ model, ED model \cite{he2018evidential} and the experiment data. It can be seen that the two models can both predict and simulate the real experimental data well. Figure \ref{compare_error} shows the error of the two models. The mean of the NQ error is less than the mean of the ED error, especially in $P_t$ and $P(A)$. Therefore, the model is considered to be able to predict successfully.

\begin{table}[htbp!]
\begin{threeparttable} 
\centering
\caption{Model results and the modified results }
\begin{tabular}{cccccccc}
\midrule
Literature & Data Type & $P(G)$ & $P(A|G)$  & $P(B)$  & $P(A|B)$ & $P_t$ &$P(A)$ \\
\hline
Busemeyer et al. \cite{busemeyer2009empirical} & Initial&  0.17 & 0.41 & 0.83 & 0.63 & 0.59 &0.69 \\
  & Result & 0.17 & 0.41 & 0.83 & 0.63 & 0.59 &0.68 \\
  & Modified result & 0.17 & 0.41 & 0.83 & 0.64 & 0.60 &0.65 \\
  Wang and Busemeyer \cite{wang2016interference} & & & & & & & \\
  Experiment 1 & Initial& 0.21&0.41&0.79&0.58&0.54&0.59\\
   &Result &0.21&0.41&0.79&0.60&0.56&0.64\\
   &Modified result &0.21&0.40&0.79&0.61&0.57&0.61\\
   Experiment 2 & Initial & 0.24&0.37&0.76&0.61&0.55&0.60\\
   & Result &0.24&0.37&0.76&0.61&0.55&0.64\\
   &Modified result &0.24&0.37&0.76&0.62&0.56&0.61\\
   Experiment 3(a) & Initial&0.24&0.33&0.76&0.66&0.58&0.62\\
   &Result&0.24&0.33&0.76&0.65&0.57&0.66\\
   &Modified result&0.24&0.33&0.76&0.66&0.58&0.62\\
   Experiment 3(b) & Initial &0.25&0.26&0.75&0.75&0.63&0.64\\
   &Result&0.24&0.33&0.76&0.65&0.57&0.66\\
   &Modified result&0.24&0.27&0.75&0.69&0.59&0.65\\
   Average & Initial&0.22&0.36&0.78&0.65&0.58&0.63\\
   &Result&0.22&0.36&0.78&0.63&0.57&0.66\\
   &Modified Result&0.22&0.36&0.78&0.65&0.58&0.63\\
\bottomrule
\end{tabular}      
\end{threeparttable}
\label{final_result}
\end{table}

\section{CONCLUSION}
\label{six}
Categorization is a significant task of many decision-making progress in knowledge-based system. 
However, a categorization disjunction fallacy is caused in some special experiments, which can be seen as an interference effect from the quantum theory perspective.
In this article, a new negation quantum decision model to predict the interference effect or disjunction fallacy, called as the proposed NQ model, is presented. 
The proposed NQ model is a combination of negation and quantum decision models. 
The proposed model tries to use these two tools to uncover more information hidden in the decision-making process. 
The general model framework is based on the quantum decision model \cite{busemeyer2011quantum}, and the exponential negation is used in the quantum probability represent from a negation perspective. 
This model has four free parameters which can be determined by a simple optimization method. 
Although the predicted results of NQ model have a general bias, they are relatively accurate after a simple modified step. 
The modified results are much more accurate compared with the existed ED model \cite{he2018evidential}.
\par
The decision model can predict this effect well in the experimental data used in this article, 
but it also has a opportunity to be used in many other applications, such as the decision support system. 
However, there also exist some open issues, for example, there was no such interference in the wider-faced, why is this? How to explain this?
In the further research, we are going to do some work on these open issues, and explore to apply this model in more application.

\section*{Acknowledgment}
The work is partially supported by National Natural Science Foundation of China (Grant No. 61973332), JSPS Invitational Fellowships for Research in Japan (Short-term).

\bibliography{myref}

\end{document}